\documentclass[11pt,a4paper]{article}
\usepackage[hyperref]{eacl2021}
\usepackage{amsmath, verbatim}
\usepackage{times}
\usepackage{sidecap}
\usepackage{floatrow}
\usepackage{latexsym}
\usepackage{graphicx}
\usepackage{footnote}
\usepackage{subcaption}
\usepackage{nccmath}

\usepackage{microtype}

\aclfinalcopy 
\def\aclpaperid{761} 

\title{Data Augmentation for Voice-Assistant NLU using BERT-based Interchangeable Rephrase}

\author{Akhila Yerukola\thanks{\enspace equal contribution}, \ Mason Bretan\footnotemark[1], \ Hongxia Jin   \\
  Samsung Research America \\
  \texttt{ \string{a.yerukola, mason.bretan, hongxia.jin\string}@samsung.com} } 

\date{}

\begin{document}
\maketitle
\begin{abstract}
We introduce a data augmentation technique based on byte pair encoding and a BERT-like self-attention model to boost performance on spoken language understanding tasks. We compare and evaluate this method with a range of augmentation techniques encompassing generative models such as VAEs and performance-boosting techniques such as synonym replacement and back-translation. We show our method performs strongly on domain and intent classification tasks for a voice assistant and in a user-study focused on utterance naturalness and semantic similarity.
\end{abstract}

\section{Introduction}

With conversational assistants becoming more and more pervasive in everyday life, task-oriented dialogue systems are rapidly evolving. These systems typically consist of Spoken Language Understanding (SLU) models, tasked with determining the domain category and the intent of user utterances at each turn of the conversation. 

The ability to quickly train such models to meet changing and evolving user needs is necessary. However, developers often find themselves in situations with access to very little labeled training data. This is especially true when new functions are deployed, and a large user base has not had a chance to utilize the function, thus, limiting the number of available utterances that can be labeled for training. Furthermore, the process of labeling large amounts of data can be time consuming and expensive. More recently, these challenges have been both enhanced and complicated by privacy concerns and legislation that may prevent the use of user utterances for training.

Much of the recent research addressing data paucity has focused on pre-training using self-supervision and vast amounts of unlabeled data \cite{devlin2018bert, radford2018improving, radford2019language}. Pre-trained models can later be fine-tuned with a much smaller amount of labeled data for specific tasks. In this work, instead of pre-training, we explore methods that enhance and expand the task-specific training set by using data augmentation. While models such as BERT prove to be both useful and relevant, we show that data augmentation during the fine-tuning stage can boost performance even on these large pre-trained models.
\textfloatsep 8pt 
\begin{table}[t]
    \centering
    \resizebox{0.9\textwidth}{!}{
    \begin{tabular}{|c|c|}
    \hline
    \textbf{Utterance} & how do I make a margherita pizza \\ \hline
    \textbf{Domain} & Recipe\\ \hline
    \textbf{Intent} & Return\_Recipe\\ \hline
    \textbf{Rephrase} & show me how to cook a margherita pizza\\ \hline \hline
    \textbf{Utterance} & can you check air quality in santa rosa \\ \hline
    \textbf{Domain} & Weather\\ \hline
    \textbf{Intent} & Air\_Quality \\ \hline
    \textbf{Rephrase} & is it possible to check air quality in santa rosa\\ \hline
    \hline
    \textbf{Utterance} & delete all emails which have come from hotel.com \\ \hline
    \textbf{Domain} & Email\\ \hline
    \textbf{Intent} & Delete\_Emails \\ \hline
    \textbf{Rephrase} & get rid of all emails which have come from hotel.com\\ \hline
    \end{tabular}
    }
    \caption{An illustrative example of a generated rephrase using our Interchangeable Rephrase, while still maintaining the original domain and intent.}
    \label{tab:sample_example}
    \vspace{-2mm}
\end{table}
We implement and compare several pre-existing techniques for data augmentation on Natural Language Understanding (NLU) tasks such as domain classification (DC) and intent classification (IC) for a voice assistant. We also introduce a new method of data augmentation called {\it Interchangeable Rephrase} (IR) with the goal of ``rephrasing'' an existing utterance using new language while maintaining the original intent or goal (see Table \ref{tab:sample_example}).

\section{Related Work}
\label{section:related_work}

Recurrent neural network (RNN)-based VAE generative models \cite{kingma2013auto, bowman2015generating} explicitly model properties of utterances like topic, style, and other higher-level syntactic features. The variational component helps in generating diverse text, thus, we use VAEs as a candidate for transforming and augmenting text data in our experiments.

Building upon unconditioned text generation, Conditional VAE (CVAE) \cite{sohn2015learning, hu2017toward} generates more relevant and diverse text conditioned on certain control attributes, e.g. tense, sentiment \cite{hu2017toward}, style \cite{ficler2017controlling}.  In this work, our goal is to maintain semantic similarity, therefore, we generate text by conditioning on the original intent or goal. 

\citeauthor{guu2018generating} generate novel sentences from prototypes by exploiting analogical relationships of sentences.  They have shown that generated sentences have a varied style. We extend this idea by using prototype utterances and then editing it into a new utterance or rephrase, using both the VAE and CVAE architecture, to generate augmented rephrase data. We call these models VAE-edit and CVAE-edit.


Back translation (BT) is the process of translating an utterance in a certain language to another language and then translating it back to the original language. Certain question answering (QA) models \cite{yu2018qanet} have observed that back-translation generates diverse paraphrases, while preserving the semantics of the original sentences. In our case, we use back-translations as rephrases to augment data.

Another form of simple augmentation techniques is EDA: easy data augmentation \cite{wei2019eda}. They consist of four operations: synonym replacement, random insertion, random swap and random deletion. They show a boost in text classification tasks with these operations on smaller datasets. Since the number of labeled of spoken utterances are limited, we compare to this approach for our NLU tasks. 
 
Automatic speech recognition (ASR) module, which converts audio input into text, introduces some errors in the process before feeding it into downstream NLU tasks. 
Without modifying ASR or NLU components, an utterance correction module can be used to help with denoising data \cite{freitag2018unsupervised}. The reconstructed utterances can be further used as data augmentation. 

PPDB is a well known paraphrase database consisting of automatically generated paraphrases \cite{ganitkevitch2013ppdb}. We use this database to rephrase utterances by identifying short phrases within the utterance and replacing them with a related phrase according to the database and POS (e.g. $\text{``there is a lot of"} \rightarrow \text{``there are plenty of"}$).

\section{Interchangeable Rephrase}

BERT is pre-trained using two tasks:  Masked LM (MLM) and Next Sentence Prediction (NSP). In our rephrase task, the end-objective is almost identical to the MLM training, and only this procedure is used to train our self-attention model. MLM training allows the model to predict appropriate word(s) to replace the masked token depending on the context of the rest of the phrase. Each input token corresponds to a final hidden vector that is fed into an output softmax over the vocabulary. Thus, to rephrase an utterance that has more or fewer tokens than the original, the desired number of tokens for the rephrase must be known a priori.

\begin{figure}[ht]
\includegraphics[width=7.5cm]{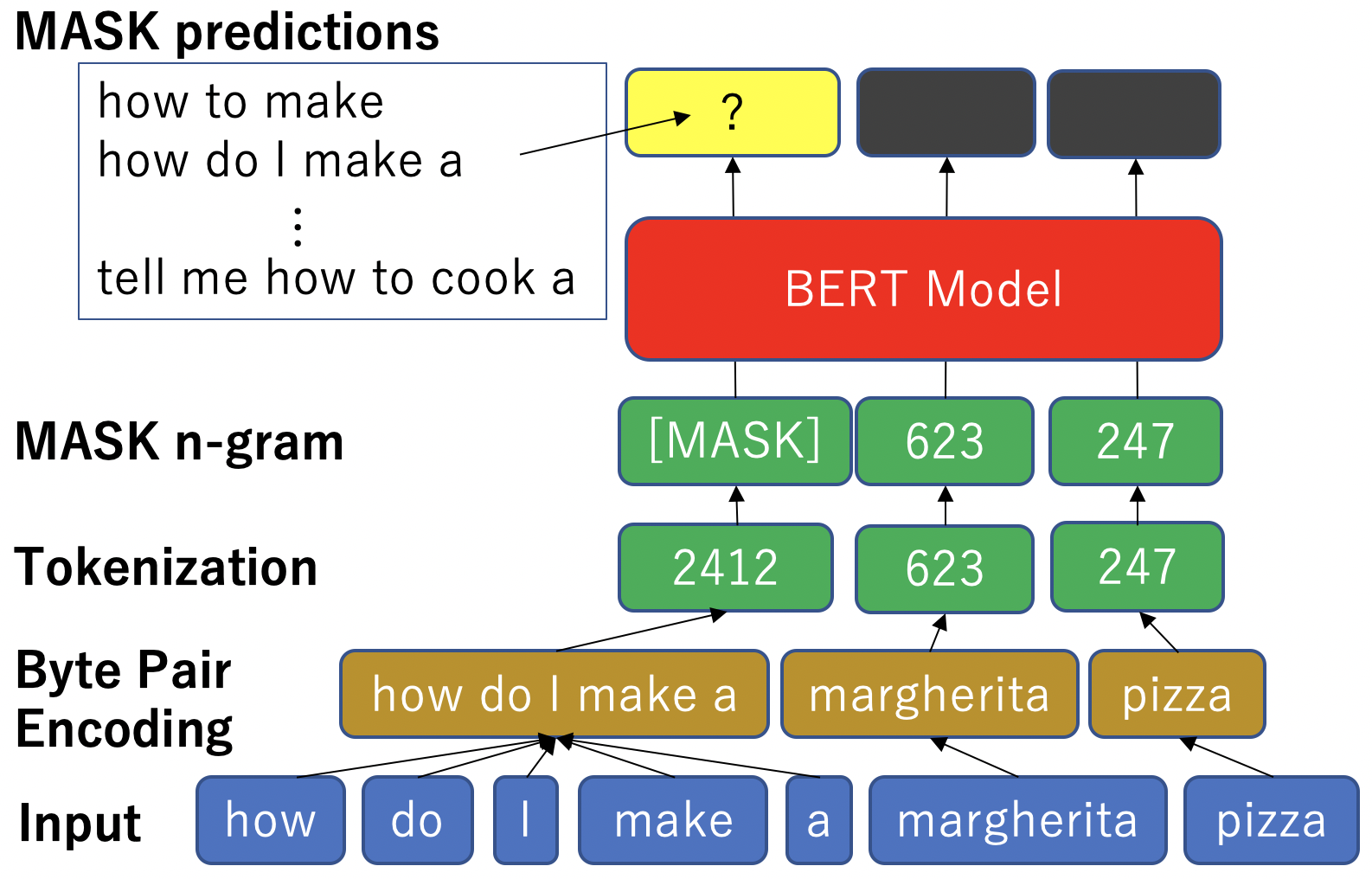}
\centering
\caption{Overview of BERT based interchangeable rephrase (BERT-IR) - BPE is used is to encode n-grams into a single token. Thus, the model's vocabulary is comprised of tokens representing both single words and sequences of words. The model then computes a softmax over the vocabulary representing the vector of the masked input token allowing for a final output that may be a different word length than the original input.}
\label{fig:rephrase_overview}
\end{figure}
To allow rephrases with unknown lengths, we use byte pair encoding (BPE) to group word $n$-grams into single tokens \cite{sennrich2015neural}. By using BPE, an individual token may represent a sequence of several words, but the model can still be trained to predict only a single token. We perform BPE on a set of training relevant to the end tasks to get the most frequent $n$-gram sequences and include these in the model's vocabulary. Similarly to PPDB, we assume that many of these $n$-grams are synonyms and are interchangeable. For example, in the context of a virtual assistant skill that enables finding and reciting recipes, $n$-gram short phrases such as a ``how to make", ``tell me how to cook" and ``teach me to make" can all be used in the place of the $n$-gram ``how do I make" in the utterance ``how do I make a margherita pizza." This interchangeable property is the foundation of our rephrase and data augmentation system.

The BERT-like self-attention model allows for the predictions to be made based on the context of the input utterance and BPE tokenization allows for variable-length outputs while still only having to predict a single token. Though the new BPE-based vocabulary requires re-training of the model (pre-trained BERT cannot be used), it remains structurally the same as the original BERT model. We refer to our BERT based interchangeable rephrase model as \textbf{BERT-IR}. Figure \ref{fig:rephrase_overview} provides an overview of the rephrase model.

In order to maintain intent such that the bigram ``turn on" in the utterance ``turn on the lights" is not inadvertently replaced with ``turn off", negative examples and an intent feature are included in a fine-tuning step. The fine-tuning process minimizes a loss function based on the cosine similarity function 
\useshortskip
\begin{equation}
\it{sim}(\vec{X}, \vec{Y}) = \frac{\vec{X}^T\cdot \vec{Y}}{|\vec{X} || \vec{Y}|}
\end{equation}
where $\vec{X}$ and $\vec{Y}$ are two equal length vectors and negative examples are included through a softmax function 
\useshortskip
\begin{equation}
\it{\it{P}(\vec{R}|\vec{Q}) = \frac{\exp(\it{sim}(\vec{Q}, \vec{R}))}{\sum_{\vec{d} \epsilon D} \exp(\it{sim}(\vec{Q}, \vec{d}))}}
\end{equation}
where  $\vec{Q}$ is a one hot ground truth token label derived from the input $Q$, and $\vec{R}$ represents the output vector of $Q$. $D$ is the set of three vectors that includes $\vec{R}$ and two vectors derived from two negative examples in the training set. The network then minimizes the following differentiable loss function using gradient descent 
\begin{equation}
\it{-log}\prod_{(Q, R)} P(\vec{R} | \vec{Q})
\end{equation}

We emphasize that it is not necessary to use a BERT model for this rephrase method. It is technically possible to use a model architecture identical to that of BERT. However, given that this is not a fine-tuning task and the model needs to be fully re-trained to support the new $n$-gram tokens a less resource intensive and data hungry model is preferred. In our experiments we use a BERT-like model that leverages self-attention, but has only a fraction of the total parameters of BERT.

\section{Experiments}
In the following experiments, we examine and compare our proposed method with various data augmentation techniques in the context of utterance generation (rephrase) for voice assistants. 

First, we study the properties of the utterances the systems are capable of generating. An ideal data augmentation method should create data that either expands upon or fills in missing gaps of the original training distribution, while still being inherently natural and meaningful. In this context, given an input utterance (from the original distribution), the goal is to generate rephrases that are semantically similar, yet, different enough to positively alter the original training distribution. This is measured in our next two experiments in which we compare the performance of augmented datasets by training an utterance domain classifier (DC) and an intent classifier (IC).

Finally, we perform a user study to examine the quality of utterance generation in terms of naturalness and semantic similarity.  

{\bf Data.} We use an original dataset comprised of utterances for a set of 63 skills (domains) for a voice assistant. The skills range from playing a song on Spotify to turning on/off in-home appliances to providing the weather. The developer of each skill provides a set of training utterances that users can say as an entry point to the function. Each utterance is annotated with an intent, and on average there are 530 utterances per skill (maximum 2000 and minimum 9). 

For the DC task, of the 43540 utterances in the dataset, 6590 are held out for validation and the remaining utterances are used for training and processed for augmentation. Additionally, we use a separate test set that was collected and labeled through user trials and crowd sourcing. Each domain in this test set has roughly 900 test utterances. 

For the IC task, we consider 5 different skills (domains) from the above dataset: Weather, SystemApp, SmartThings, TvControl and TvSettings. We allocate 30\% of the data in each skill for testing and use the remaining for training including augmentation. (see Appendix \ref{section:app-datastats} for more details). 

\begin{figure*}[th]
    \centering
    \captionsetup[subfigure]{oneside,margin={1.5cm,0cm}}
    \hskip -0.6in
    \begin{subfigure}[t]{0.29\textwidth}
        \centering
        \includegraphics[ scale=0.19]{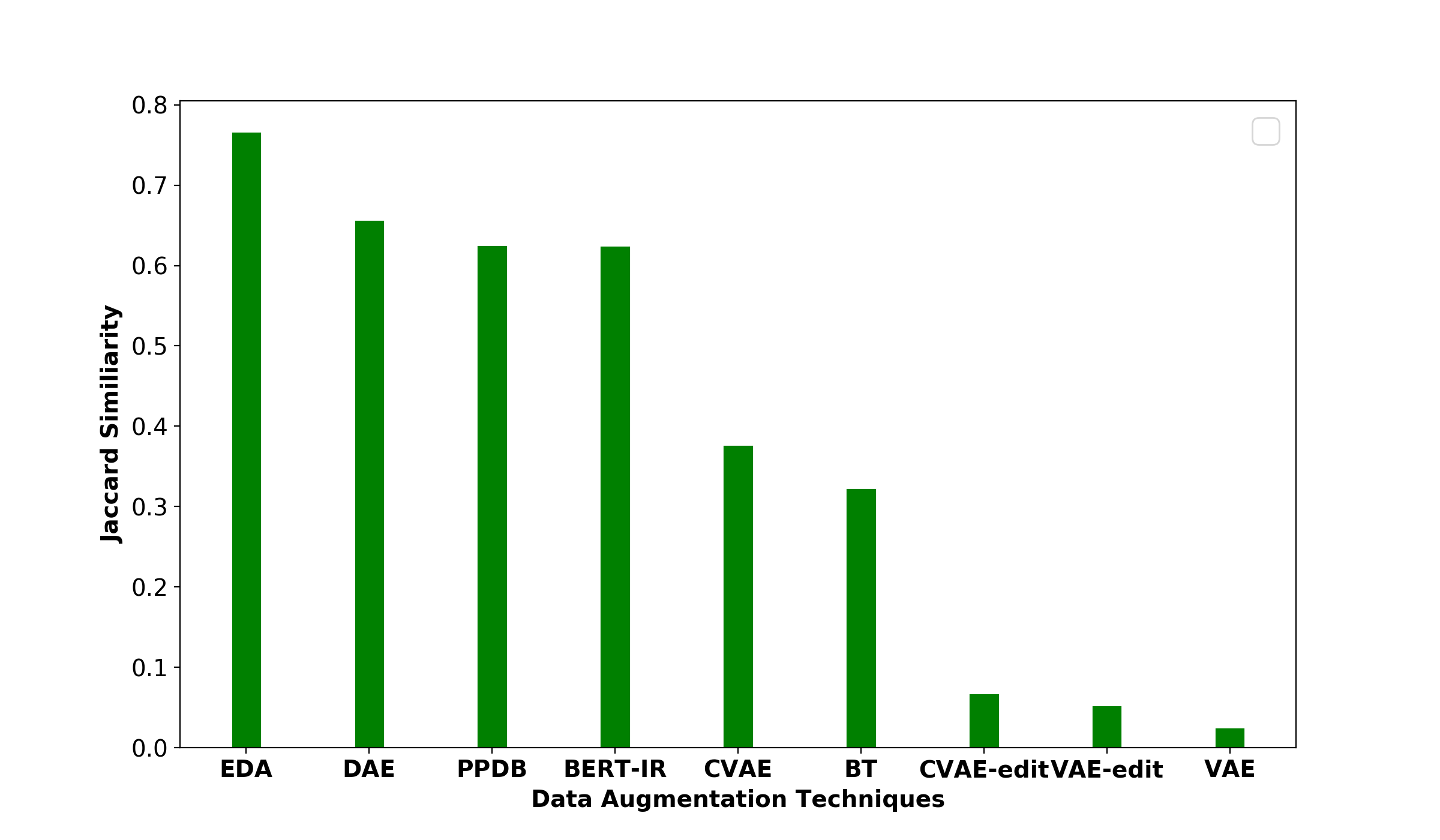}
        \caption{\textbf{Jaccard Similarity:} \\   $\downarrow$ lower is better}
    \end{subfigure}
    \hspace{1em}
    \begin{subfigure}[t]{0.32\textwidth}
        \centering 
        \includegraphics[scale=0.166]{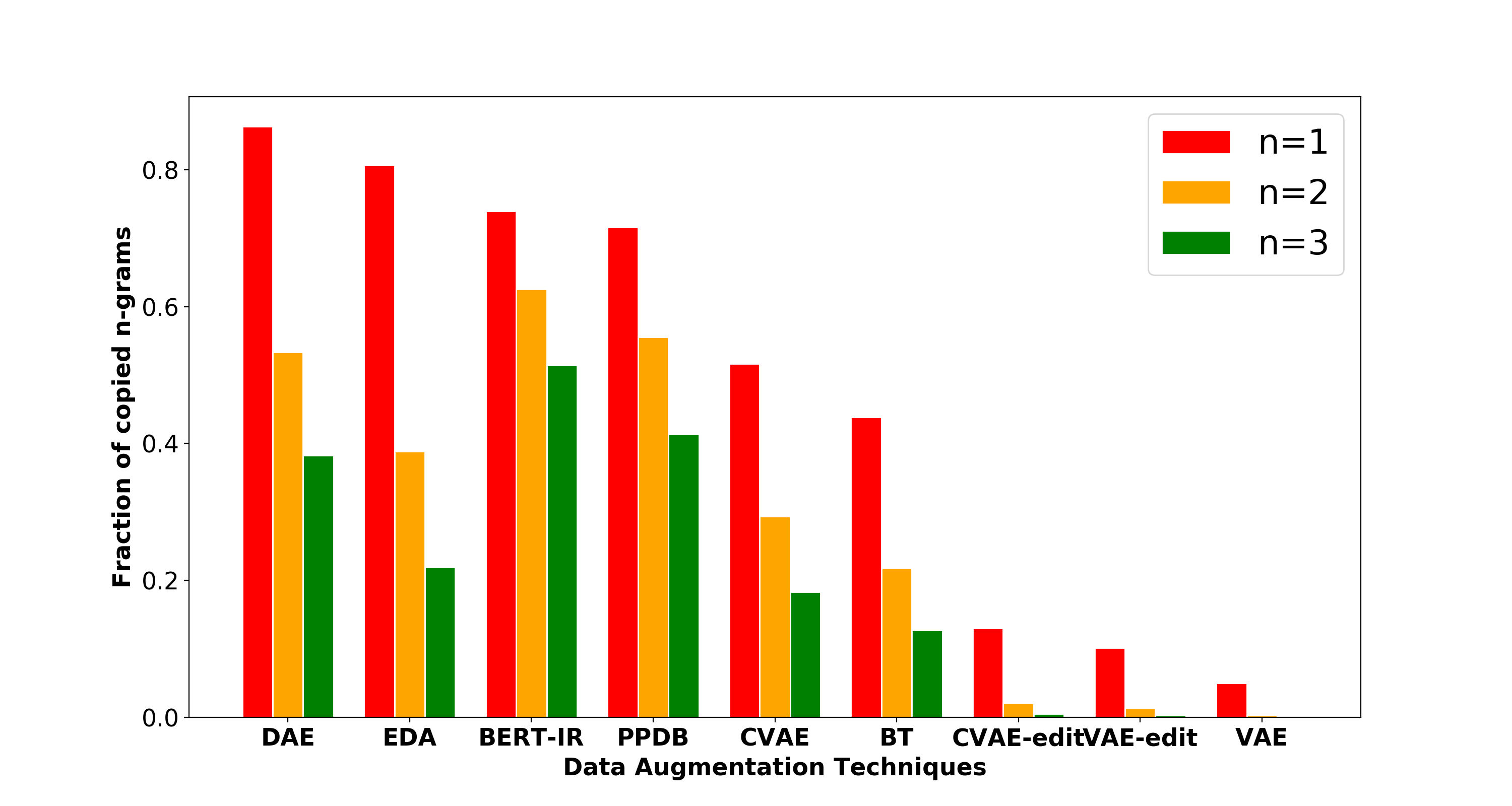}
        \caption{\textbf{Copied $n$-gram fraction:}  $\downarrow$ lower is better}
    \end{subfigure}
    \hspace{1em}
    \begin{subfigure}[t]{0.3\textwidth}
        \centering
        \includegraphics[scale=0.166]{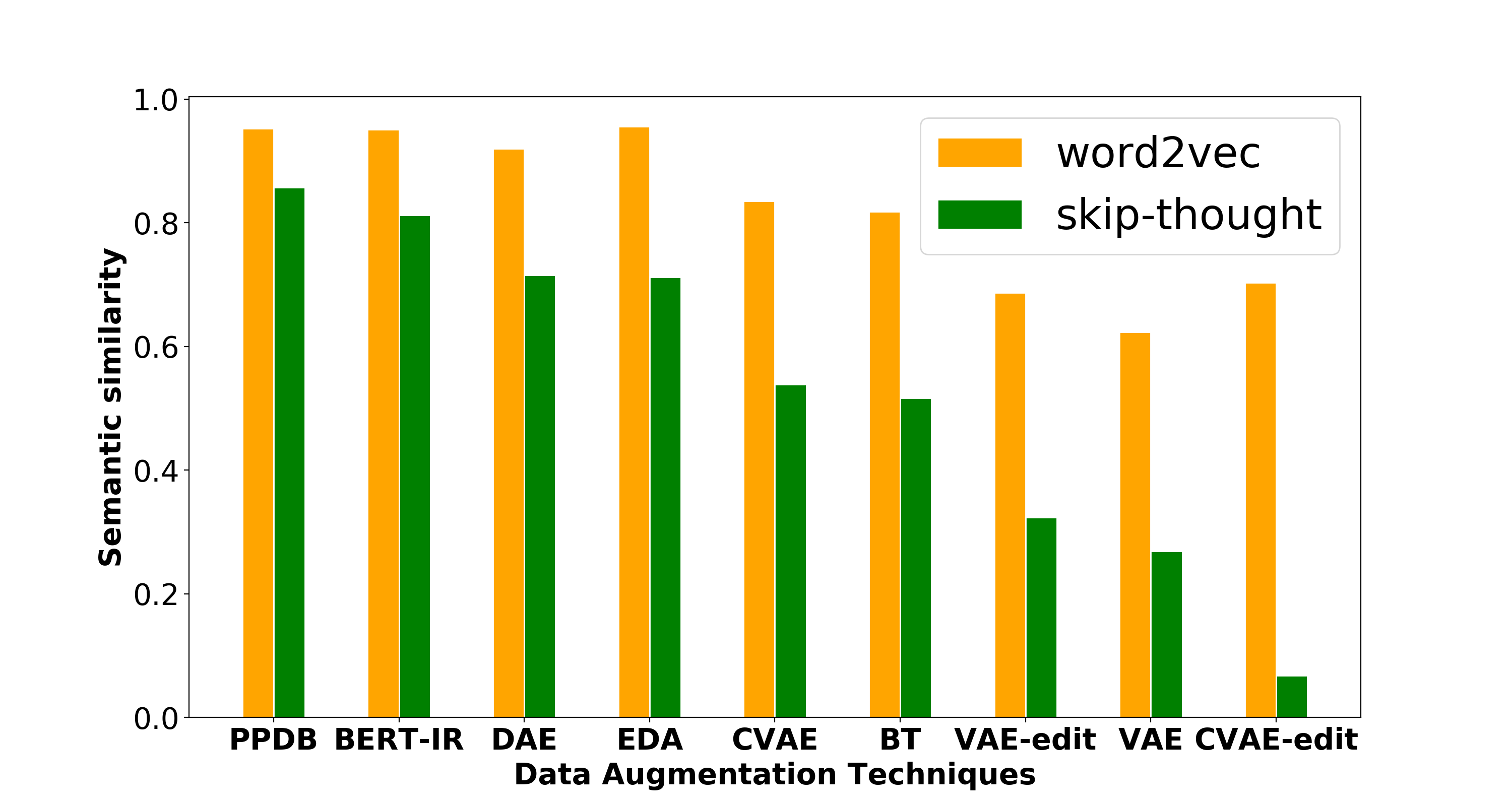}
        \caption{\textbf{Semantic Similarity:} \\ $\uparrow$ higher is better}
    \end{subfigure}
    \caption{Evaluation of automatic metrics: Averaged across all the generated rephrases for each model.}
    \label{fig:auto_metrics}
\end{figure*}

\subsection{Experimental Settings}
\label{sec:exp_settings}
Our BERT-based interchangeable rephrase model uses an architecture that is similar in size and structure to DistilBERT \cite{sanh2019distilbert}. It is possible to use alternative types of models, but we are motivated by the power of self-attention mechanisms for sequential tasks. Our model is pre-trained with roughly 500k utterances from user data (in USA) for the voice assistant. The vocabulary is established using a combination of the developer training data and the usage data. After the byte-pair-encoding process, we prune the resulting pairs so that only pairs occurring more than 100 times are used in the final vocabulary. All unigrams appearing two or more times are also included in the vocabulary.

{\bf Baselines.} Our first baseline is a classifier that is finetuned solely on the original training data with no augmentation. We compare with existing work on data augmentation via VAE/CVAE, VAE/CVAE-edit, back-translation, denoising autoencoder (DAE), easy data augmentation (EDA) and PPDB. We refer the reader to Appendix \ref{section:app-expdetails} for experimental details of the above augmentation techniques.

\subsection{Analysis and Comparison of Methods}
\label{sec:automatic_metrics}
We apply several automated linguistic metrics to examine differences in quality of generated rephrases. We evaluate the generated rephrases based on how \textit{related} they are to the original utterances. A model which minimizes the word-level overlap (i.e more variation) and increases the semantic similarity the most is presumably ideal.

Jaccard similarity \cite{jaccard1912distribution, roemmele2017evaluating} is used to measure the proportion of overlapping words between the rephrase and the original utterance. Additionally, for $n = 1, 2, 3$, we measure the proportion of generated $n$-grams that also appear in the original utterance i.e amount copied from original utterance \cite{see2019massively}.  

We measure semantic similarity at the word-level and sentence-level. We compute the the mean cosine similarity of the word2vec vectors of all pairs of words between a rephrase and the original utterance. We also measure the cosine similarity of the sentence encodings, generated by the skip-thought model \cite{kiros2015skip}, of the rephrase and the original utterance (see Figure \ref{fig:auto_metrics}c). The skip-thought model maps sentences sharing semantic and syntactic properties to similar vector representations. 

We average all these metrics across all the generated rephrases for each model (see Figure \ref{fig:auto_metrics}).

\begin{table}[t]

\begin{center}
\begin{small}
\begin{sc}
\begin{tabular}{ccc}
    \hline 
    \textbf{Model} & \textbf{Held Out} & \textbf{Test Acc} \\ \hline
    No - augmentation & 0.9229 & 0.8251 \\ 
    VAE & 0.9212 & 0.8379 \\ 
    CVAE & 0.9277 & 0.8447 \\ 
    VAE-edit & 0.84 & 0.6891 \\ 
    CVAE-edit & 0.8617 & 0.7469 \\ 
    Back-translation & 0.9361	& 0.8881 \\
    DAE & 0.9307 & 0.8765 \\
    EDA & 0.9896 & 0.8921 \\
    PPDB & 0.9341 & 0.8764 \\ 
    BERT-IR & 0.9738 & {\bf 0.9062} \\ 
    \hline
    
    \end{tabular}
\caption{Domain classification results.  }
\label{tab:cc}
\end{sc}
\end{small}
\end{center}
\end{table}

\subsection{Domain and Intent Classification}

\begin{SCfigure}[2.2][th]
\includegraphics[width=0.17\textwidth]{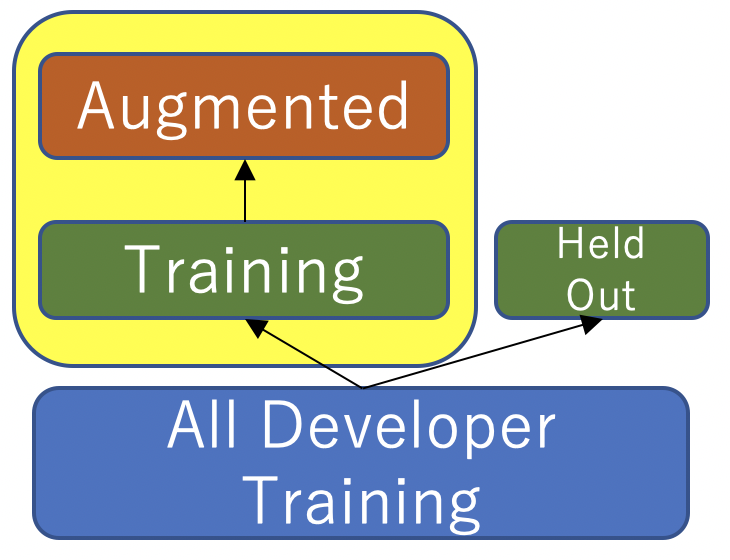}
\centering
\caption{Training distribution for domain and intent classification. All data within yellow boundary box is used for training the classifier.}
\label{fig:classifier_dist}
\end{SCfigure}

The data generated from each augmentation technique is used to train a domain classifier and an intent classifier. Ten classifiers are trained for each task, using the same distribution of training and held out data described previously. Though several of the augmentation techniques are capable of generating more than one utterance per input, here we generate augmented data in a 1-1 fashion i.e, for each original utterance in the training set, a single rephrase is generated (see Figure \ref{fig:classifier_dist}). 

\begin{figure}[t]
\includegraphics[width=1.1\textwidth]{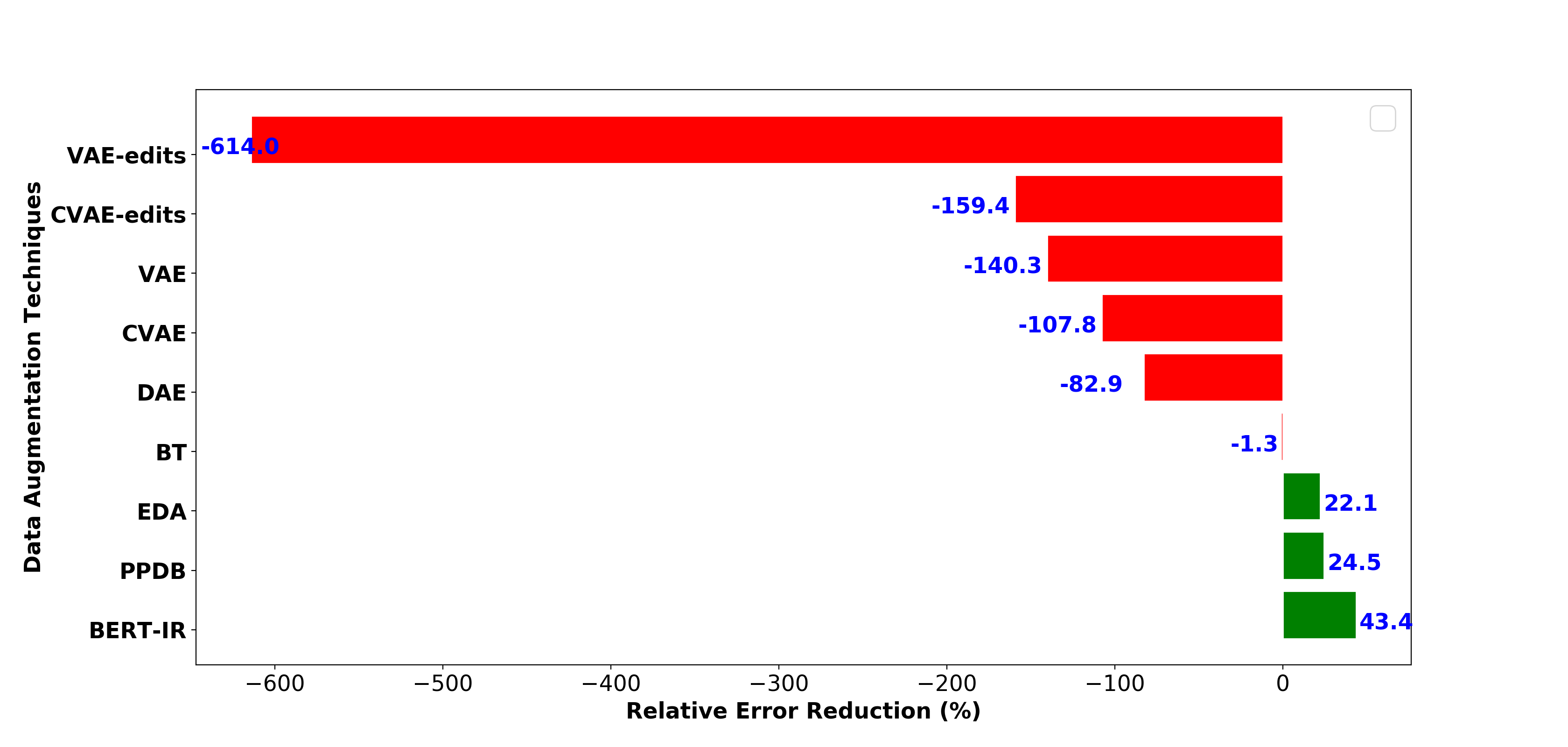}
\centering
\caption{Intent Classification: Relative error reduction of each augmentation type compared to the original ``no augmentation'' classifier. A positive value (green bar) indicates better performance.}

\label{fig:intent_classification}
\end{figure}

For both the tasks, we train a classifier on top of the base uncased DistilBERT \footnote{\url{https://huggingface.co/transformers/model_doc/distilbert.html}} model (see  \ref{section:ic-dc}). 
The results on DC task are shown in Table \ref{tab:cc}. In Figure \ref{fig:intent_classification}, we show the relative error reduction of each augmentation technique achieved on IC, averaged across all 5 skills (see Appendix \ref{section:ic_results_sec} for complete details).

\subsection{User study}
We performed an online user study where participants completed two tasks. In the first task, participants answered a three item Likert survey regarding the naturalness of the utterance (see Appendix \ref{section:user-study}). We compared results for utterances that came from the original developer training (i.e. human-generated), our proposed BERT-IR, and a baseline rephrase using PPDB.

In the second task, users were given a reference utterance and four candidate utterances. Of the four candidates, one of them had an identical intent as the reference, but with different wording, and the other three were random utterances with different intents. The participants were asked to select the candidate with the same intent as the reference. The ``correct" candidate was generated from one of the three sources: human, our rephrase, or PPDB rephrase. There were 88 participants, and each participant completed each task 30 times. Results for both tasks are shown in Table \ref{tab:user_study}. Using a 3-way ANOVA, we found significant differences between all three methods on a $p<.05$ level.

\begin{table}[t]
\begin{center}
\begin{small}
\begin{sc}
    \begin{tabular}{ccc}
    \hline 
         \textbf{Source} & \textbf{Avg. Naturalness} & \textbf{Intent Accuracy}\\ \hline
         Human & 3.9 & 84.5 \\ 
         PPDB & 3.1 & 72.3 \\ 
         Ours & 3.4 & 78.6 \\
         \hline
    \end{tabular}
    \caption{User study results. The average likert naturalness (where 5 is very natural and 1 is very unnatural) and intent matching accuracy are reported for each of the three utterance sources.}
    \label{tab:user_study}
\end{sc}
\end{small}
\end{center}
\end{table}

\section{Discussion}
In our experiments, we show that VAE models are capable of generating diverse rephrases. However, these rephrases do not preserve the original meaning. This likely contributed to poorer performance on DC and IC. The discriminator of CVAE models trained to condition on a domain (DC) or intent (IC) helps improve semantic similarity resulting in slightly improved performance of NLU tasks. The VAE-edit and CVAE-edit models perform quite poorly on all comparisons as they don't necessarily preserve the meaning of the utterance when transforming them into an altered style.

Back-translation yields relatively diverse rephrases; however, it poorly conserves the original meaning. DAE just tends to copy a higher fraction of $n$-grams, while not changing the meaning of the utterance. EDA has boosted the DC and IC performance, compared to other methods. However, it merely changes a small percent of words in utterances by replacement operations, which are not always grammatically sound.

Our BERT-IR yields the best performance on the NLU tasks, with a relative error reduction of 46.26\% on DC and 43.4\% on IC as compared to no additional augmentation. 
Similar to PPDB, our approach generates rephrases with a lower word-overlap and a significantly higher semantic similarity with the original utterance. 
The user study reveals that BERT-rephrase is an improvement over the PPDB baseline, but does not perform at the level of human generated utterances in terms of naturalness and intelligibility. 

For examples of generated rephrases see Table \ref{tab:examples} in the Appendix.
\section{Conclusion}
We introduced BERT-IR, a simple augmentation strategy based on byte pair encoding and a BERT-like self-attention model to generate diverse, natural and meaningful rephrases in the context of utterance generation for voice assistants. We demonstrate that BERT-IR performs strongly on spoken understanding tasks like domain classification and intent classification and in a user-study focused on evaluating quality of rephrases based on naturalness and interpretability (intent preservation).



\bibliographystyle{acl_natbib}
\bibliography{eacl2021}

\clearpage

\appendix

\section{Appendices}
\label{sec:appendix}

\subsection{Implementation and Training Details of Augmentation Strategies}
\label{section:app-expdetails}

\textbf{VAE:} The decoder and encoder of the VAE model are set as single-layer GRU RNNs with input/hidden dimension of 100/150 and max sample length of 15. Larger input/hidden dimension models performed similarly. To avoid vanishingly small KL term in the VAE module, we use a KL term weight linearly annealing from 0 to 0.15 during training (similar to training procedure in \citeauthor{hu2017toward}). Training a larger VAE model yielded comparable results.

\textbf{CVAE:} The discriminator is set as ConvNets similar to \citeauthor{hu2017toward}. Balancing parameters are set to $\lambda_c  = \lambda_z = \lambda_u = 0.1$. Training procedure is exactly as \citeauthor{hu2017toward}.

\textbf{VAE edit, CVAE-edit:} We match tuples of utterances which have the same domain and intent. These pairs are used as the prototypes and new utterances during training of VAE-edit and CVAE-edit (similar to training procedure in \citeauthor{guu2018generating} by constructing edit vectors and concatenating it with $z$ before feeding into the decoder).

\textbf{Denoising Auto-encoder:} The decoder and encoder of the denoising auto-encoder are set as single-layer GRU RNNs with input/hidden dimension of 300/512 with Luong attention. We follow a similar utterance corruption as \citeauthor{freitag2018unsupervised} by dropping words randomly whose frequency was greater than 100 (i.e non-content words like "the"), followed by shuffling of its bigrams while not splitting bigrams that also exist in the original utterance. We train a DAE using this corrupted data. We use the corrected utterance/rephrases generated as augmentation data. 

We train the above models until convergence on 2 NVIDIA Tesla V100 GPUs.

\textbf{Back-translation:} We use the open-sourced back-translation system from \cite{xie2019unsupervised}. Specifically, we use pre-trained WMT’14 English-French translation models (in both directions) to perform back-translation on each utterance.


\textbf{EDA:} We followed three strategies by \citeauthor{wei2019eda}: random insertion, random replacement and random swap. 

\textbf{PPDB:} The PPDB \footnote{\url{http://paraphrase.org/\#/download"}} database consists of 1 paraphrase rule per line. A standard format of a line is: 
\begin{verbatim} LHS ||| PHRASE ||| PARAPHRASE ||| 
(FEATURE=VALUE )* ||| ALIGNMENT ||| 
ENTAILMENT
\end{verbatim}. 
If a PHRASE exists in the utterance, we replace it with the PARAPHRASE. Since it's a 1-1 data augmentation, we sample a rephrase from the list of all possible generated rephrases. We use the English-Phrasal PPDB database .


Refer to Table \ref{tab:inference_time} for the average run time for the different augmentations techniques.
\begin{table}[ht]
    \begin{center}
\begin{small}
\begin{sc}
    \begin{tabular}{|c|c|}
    \hline 
         \textbf{Model} & \textbf{Time}\\ \hline\hline
         VAE (GPU) & 1 minute \\ \hline
         CVAE (GPU) & 1.5 minutes\\ \hline
         VAE-edit (GPU) & 2 minutes\\ \hline
         CVAE-edit (GPU) & 3 minutes \\ \hline
         Back-translation (GPU) & 30 minutes\\ \hline
         DAE (GPU) & 1 minute\\ \hline
         EDA (CPU) & 3 minutes \\ \hline
         PPDB (CPU) & 11 minutes \\ \hline
         BERT-Rephrase (GPU) & 7 minutes\\ \hline
    \end{tabular}
    \end{sc}
\end{small}
\end{center}
    \caption{Average time to run inference for augmentation over the training set (36,950 utterances) assuming one generated output per utterance.}
    \label{tab:inference_time}
\end{table}

\subsection{Domain Classification and Intent Classification}
\label{section:ic-dc}
For domain classification and intent classification, we train a classifier on top of the base uncased DistilBERT model. We use a maximum sequence length of 128, a dropout rate of 0.1, and a learning rate of 2e-5. We train the classifier for 15 epochs and batch size of 32 on a single NVIDIA Tesla V100. 
\begin{table}[h]
    \begin{center}
\begin{small}
\begin{sc}
    \begin{tabular}{|c|c|}
    \hline 
         \textbf{Splits} & \textbf{Num. of utterances} \\ \hline
         Train & 36950 \\ \hline
         Held Out & 6590 \\ \hline
         Test & 56820 \\ \hline
    \end{tabular}
    \end{sc}
\end{small}
\end{center}
    \caption{Domain Classification Data Statistics}
    \label{tab:dc_data_stats}
\end{table}
\begin{table*}[]
    \begin{center}
\begin{small}
\begin{sc}
    
    \begin{tabular}{|c|c|c|c|}
    \hline 
         \textbf{Domain/Skill} & \textbf{Num. of intents} & \textbf{Train data size} & \textbf{Test data size} \\ \hline
         Weather &22 & 732 & 313 \\ \hline
         SystemApp &29 & 665 & 286\\ \hline
         SmartThings &61 & 1190& 511\\ \hline
         TvControl &19 & 392&168\\ \hline
         TvSettings &40 &834& 357\\ \hline
    \end{tabular}
    \end{sc}
\end{small}
\end{center}
    
    \caption{Intent Classification Data Statistics}
    \label{tab:ic_data_stats}
\end{table*}

\subsection{Training Data Statistics}
\label{section:app-datastats}
\textbf{Domain Classification:} Table \ref{tab:dc_data_stats} shows the counts for each training, development (held out), and test set. The partition used for training is also the partition that is augmented (and subsequently also used for training in domain classification, see Figure \ref{fig:classifier_dist}). Each utterance has an average of 6 words. There are 63 different domains in our dataset like music, calendar, calculator, weather etc. We evaluate the automatic metrics from Section \ref{sec:automatic_metrics} on the same data used for training the domain classifier.

\textbf{Intent Classification:} Table \ref{tab:ic_data_stats} shows the training split counts for each of the 5 different domains. Similar to domain classification, the partition used for training is also the partition that is augmented (and subsequently also used for training in intent classification, see Figure \ref{fig:classifier_dist}).

\subsection{Intent Classification Results}
\label{section:ic_results_sec}
 As shown in Table \ref{tab:ic_data_stats}, the amount of training data available to train an intent classifier for each domain is very less. This can explain the poor performance of VAE models and its variations on IC since they require a lot more training data to perform well. This is also illustrated in the results of domain classification where these VAE based models achieve a slight improvement since there is a higher amount of training data available.  Please refer to Table \ref{tab:ic_results} for the fine-grained performance results of the augmentation techniques on the 5 domains.

\begin{table*}[]

\begin{center}
\begin{small}
\begin{sc}
\begin{tabular}{cccccc}
    \hline 
    \textbf{Model} & \textbf{Weather} & \textbf{SystemApp} & \textbf{SmartThings} & \textbf{TvControl} & \textbf{TvSettings} \\ \hline
    No - augmentation  & 0.97129&	0.97894&	0.95882	&0.92814&	0.92977 \\ 
    VAE & 0.92971&	0.91929&	0.92745&	0.88023&	0.88764 \\ 
    CVAE & 0.93062&	0.95087&	0.92941&	0.88023&	0.88764 \\ 
    VAE-edit & 0.72966	&0.81403&	0.68823	&0.80239&	0.9382 \\ 
    CVAE-edit & 0.92344	&0.9614&	0.87058&	0.88622&	0.87921 \\ 
    Back-translation & 0.97368&	0.97192&	0.95294&	0.93413&	0.95505 \\
    DAE & 0.94976&	0.95438	&0.92352&	0.8982&	0.91853 \\
    EDA &0.99361&	0.98245&	0.96078&	0.93413&	0.92696 \\
    PPDB & 0.98564&	0.98245&	0.9666&	0.93413	&0.93539 \\ 
    BERT-IR & \textbf{0.9968}&	\textbf{0.98596}&	\textbf{0.97058}	&\textbf{0.94011}&	\textbf{0.94382} \\ 
    \hline
    
    \end{tabular}
\caption{Test accuracy for Intent Classification}
\label{tab:ic_results}
\end{sc}
\end{small}
\end{center}

\end{table*}

\subsection{User Study Details}
\label{section:user-study}
Participants responded to three items on a Likert-scale of one to five:
\begin{enumerate} 
    \item This utterance is natural. (disagree to agree)
    \item It is easy to understand the intent. (disagree to agree)
    \item A person is likely to say this to a voice assistant to achieve the underlying goal. (disagree to agree)
\end{enumerate}

\begin{table*}[t]
    \centering
    \begin{tabular}{c|c}
        
        \textbf{Model} & \textbf{Generated Rephrase} \\ \hline\hline
        	Original & \textit{barbershop near me} \\ \hline 
            VAE &	send an email to tony \\ \hline 
            CVAE &	barbershop near me \\ \hline 
            VAE-edit &	is mike's bikes closed on weekends? \\ \hline 
            CVAE-edit &	where is the highest rated dentist nearby? \\ \hline 
            Back-translation &	the "Teller" Argile been in close commute for us! \\ \hline 
            DAE &	barbershop near near me \\ \hline 
            EDA	& near me barbershop \\ \hline 
            PPDB &	barbershop near me \\ \hline 
            BERT-IR &	barbershop close to my current location \\ \\
    \end{tabular}
    \begin{tabular}{c|c}
        
        \textbf{Model} & \textbf{Generated Rephrase} \\ \hline\hline
        	Original & \textit{ call an uber espanol to los altos taqueria for me} \\ \hline 
            VAE &	what is the time difference between kuala lumpur and and 40 feet \\ \hline 
            CVAE &	call holly davis espanol to a stadium in brooklyn \\ \hline 
            VAE-edit &	order a lyft premier to 1 \\ \hline 
            CVAE-edit &	order a lyft lux for 2 to a cake for me \\ \hline 
            Back-translation &	call a espanol uber at altos taqueria commands for me \\ \hline 
            DAE & look for an espanol espanol los altos \\ \hline 
            EDA &	call an uber espanol to los for me altos taqueria \\ \hline 
            PPDB &	'm calling the uber espanol to los altos taqueria for me \\ \hline 
            BERT-IR &	please contact an uber espanol to los altos taqueria for me \\ \\
            
    \end{tabular}

    \begin{tabular}{c|c}
        
        \textbf{Model} & \textbf{Generated Rephrase} \\ \hline\hline
        	Original & \textit{how is today's visibility?} \\ \hline 
            VAE &	send a lyft premier to see you \\ \hline 
            CVAE &	how is today's visibility? \\ \hline 
            VAE-edit & what's the weather going to be here \\ \hline 
            CVAE-edit &	navigate to golden gate bridge \\ \hline 
            Back-translation & How do we make today's visibility? \\ \hline 
            DAE &	what is today's visibility? \\ \hline 
            EDA &	how is nowadays's visibility? \\ \hline 
            PPDB &	how is today's visibility? \\ \hline 
            BERT-IR &	i want to know how's today's visibility
            
    \end{tabular}
    \caption{Examples of generated rephrases by the different augmentations techniques}
    \label{tab:examples}
\end{table*}

\end{document}